\documentclass[twoside]{article}

\usepackage[accepted]{aistats2017}

\usepackage{amsfonts,eucal,amsbsy,amsthm,amsopn,amssymb,amsmath}
\usepackage{enumitem}
\usepackage{graphicx}
\usepackage{booktabs}
\usepackage{tikz}
\usetikzlibrary{matrix}
\usetikzlibrary{arrows}
\usepackage{pgfplots}
\usepackage{hyperref}
\newtheorem{definition}{Definition}

\usepackage{natbib}

\newtheorem{property}{Property}
\newtheorem{theorem}{Theorem}
\newtheorem{lemma}{Lemma}
\newcommand{\BigO}[1]{\ensuremath{\operatorname{\mathcal{O}}\bigl(#1\bigr)}}

\newcommand{\ignore}[1]{}
\newcommand{\rr}{\mathbb{R}}
\newcommand{\mat}[1]{\mathbf{#1}}
\newcommand{\Ps}{\mathcal{P}}
\newcommand{\Ss}{\mathcal{S}}
\renewcommand{\H}{\mathbf{H}}
\newcommand{\A}{\mathbf{A}}
\newcommand{\azero}{\boldsymbol{\alpha}_0}
\newcommand{\ainf}{\boldsymbol{\alpha}_\infty}
\newcommand{\Train}{\mathcal{T}}

\begin{document}

%
\runningtitle{Maximum Bipartite Matching for Spectral Learning}

%

\twocolumn[

\aistatstitle{A Maximum Matching Algorithm for Basis Selection \\in Spectral Learning}

\aistatsauthor{ Ariadna Quattoni \and Xavier Carreras \and Matthias Gall{\'e} }

\aistatsaddress{ Xerox Research Centre Europe (XRCE) \\ Meylan, France\\ {\small \tt \{ariadna.quattoni,xavier.carreras,matthias.galle\}@xrce.xerox.com }} ]

\begin{abstract}
  We present a solution to scale spectral algorithms for learning
  sequence functions.  We are interested in the case where these
  functions are sparse (that is, for most sequences they return $0$).
  Spectral algorithms reduce the learning problem to the task of
  computing an SVD decomposition over a special type of matrix called
  the Hankel matrix. This matrix is designed to capture the relevant
  statistics of the training sequences. What is crucial is that to
  capture long range dependencies we must consider very large Hankel
  matrices. Thus the computation of the SVD becomes a critical
  bottleneck. Our solution finds a subset of rows and columns of the
  Hankel that realizes a compact and informative Hankel submatrix.
  The novelty lies in the way that this subset is selected: we exploit
  a maximal bipartite matching combinatorial algorithm to look for a
  sub-block with full \textit{structural} rank, and show how
  computation of this sub-block can be further improved by exploiting
  the specific structure of Hankel matrices.
\end{abstract}

\section{INTRODUCTION}

Our goal is to model functions whose domain are discrete sequences
over some finite alphabet. Our focus is on sparse functions, by which
we mean functions that have the property that only a very small
proportion of the sequences in the domain map to a non-zero value.
We call those sequences the support of the function.
The main motivation lies in solving problems arising in Natural
Language Processing (NLP) applications, where sparse sequence
functions are of special interest.  For example, think of all possible
sequences of $T$ letters that constitute valid English words of length
$T$.  If $\Sigma$ is the set of English letters, is clear that out of
the $\Sigma^T$ possible letter sequences only a very small fraction
are valid words (i.e. should have non-zero probability).

One interesting function class over $\Sigma^\star$ is that of
functions computed by Non-Deterministic Weighted Automata (WA), since
this class properly includes classes such as ngram models and hidden
Markov models.
In recent years several approaches for estimating WAs have been
proposed that are based on representing the function computed by a
WA using a Hankel matrix \citep{multiplicity,jaeger,hsu09,moments,mlj12}.

As an illustration of the method, consider the following problem:
Assume we are given a set of pairs $(x,f(x))$, where $x$ is a sequence
in the support of some target function $f$ over $\Sigma^{\star}$ and we wish to
learn a WA that approximates $f$.
The spectral method provides a solution to this problem and it would
work in four steps:
\begin{enumerate}
\item Basis Selection: Choose a set of prefixes $\Ps$ and suffixes $\Ss$.
\item Build a Hankel matrix: $\H \in \rr^{|\Ps| \times |\Ss|}$ where the entry $\H(p,s)$ is the value of the target
  function on the sequence obtained by concatenating prefix $p$ with suffix $s$.
\item Perform SVD on $\H=\mat{U}  \mat{\Sigma} \mat{V}^\top$.
\item Use the factorization $\mat{F}=\mat{U} \mat{\Sigma} $ and
  $\mat{B} = \mat{V}^\top$ and $\H$ to recover the parameters of the
  minimal WA, following \citet{hsu09} (see \S\ref{section:spectral} for details).
\end{enumerate}

The computational cost of the algorithm will be dominated by the SVD
step $~\BigO{\min(|\Ps|,|\Ss|)^3}$, thus to control the computational
complexity, it is critical to choose a small and yet informative basis.

The theory of spectral learning tells us that if the target function
has a minimal WA representation of size $n$, there will be a complete
basis where $|\Ps|=|\Ss|=n$, where complete means that the rank of the
corresponding Hankel defined over that basis is the same as the size
of the minimal WA. But the theory does not give a practical answer to
how to choose such a basis. The design of efficient algorithms for
choosing an informative and yet small sample-dependent basis is still
an open problem, which is the focus of our paper.

We propose an efficient combinatorial algorithm for sample-dependent
basis selection.
At its core, our strategy computes a maximum matching of the bipartite
graph associated with the sparsity pattern of a Hankel matrix.
%
The main idea is quite simple, we find a subset of
prefixes and suffixes in the given sample, such that the corresponding
Hankel matrix defined over that basis has full structural rank. The
key insight is that for sparse matrices it is easy to remove symbolic
dependencies (i.e. dependencies at the level of the sparsity pattern
of the matrix). Similar ideas have a long history in the numerical
optimization literature, where combinatorial algorithms are used for
computing preconditioners for solving large sparse linear
systems. However, to the best of our knowledge we are the first ones
in applying this idea in the context of spectral learning.

We show that when the Hankel matrix of a function satisfies some
non-degeneracy assumptions, our basis selection algorithm is optimal,
in the sense that it computes the smallest complete basis. While the
non-degeneracy assumption will not be always satisfied, our
experiments suggest that it is always almost satisfied for sparse
sequence functions.

Our experiments on a real sequence modeling tasks show that the
proposed algorithm can select a basis that is at least an order of
magnitude smaller than the best alternative methods for basis
selection, resulting in an SVD step which is at least two orders of
magnitude faster.

\subsection{Related Work}

Although choosing a basis is in practice an important task for having
a robust spectral learning algorithm, not much research has focused on
this problem. One popular approach is to choose a basis by selecting
all observed prefixes and suffixes of length less than $T$, for some
$T > 0$ \citep{hsu09,Siddiqi10a}. In practice, this strategy only
works if there are no long-range dependencies in the target function.
\citet{wiewiora2005learning} presented a greedy heuristic where for
each prefix added to the basis a computation taking exponential time
in the number of states $n$ is required.  \citet{denis} suggest to
include all observed prefixes and suffixes (observed in the sample) in
the basis.
There are some theoretical results \citep{denis16} that
suggest that under certain assumptions this is the optimal strategy,
in the sense that there is no \emph{statistical harm} in considering all
prefixes and suffixes.
However, this approach is in practice unfeasible: to give a concrete
example if one considers modeling the distribution of n-grams up to
length 10 in a standard NLP benchmark, the unique number of observed
prefixes and suffixes is at least tens of millions.
Finally, \citet{balle12} gave the first theoretical results for the
problem of basis selection. They show that by sampling prefixes and
suffixes proportional to their frequency in a large enough sample,
with high probability, a complete basis will be found. They also
provide experimental results \citep{mlj12}.

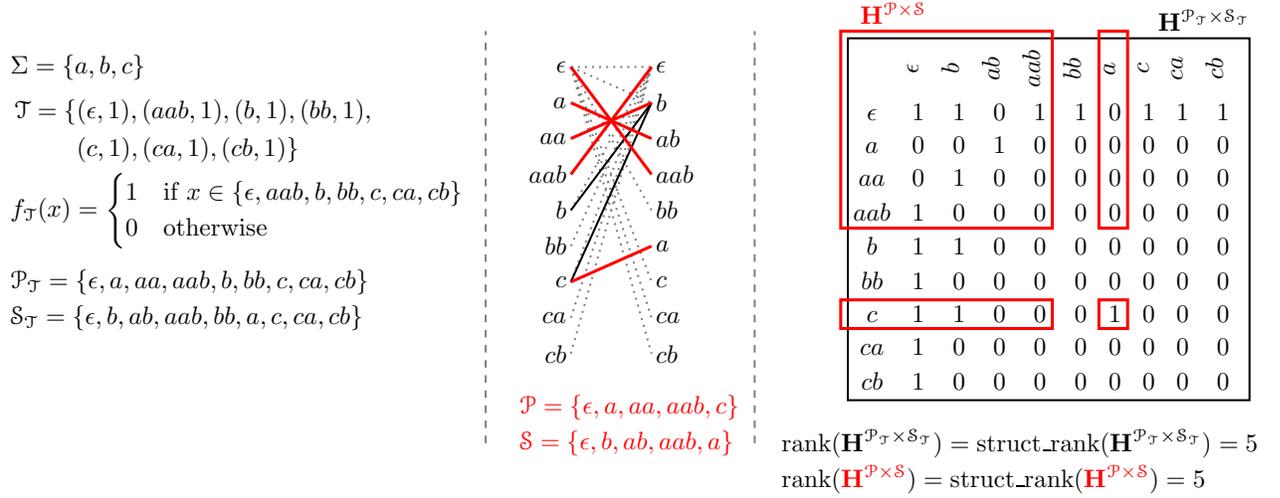
\begin{figure*}[t]
  \resizebox{\linewidth}{!}{
%
%
%
%
%

\begin{tikzpicture}[scale=0.5]

  \node [left] at (0,9) {$\epsilon$};
  \node [left] at (0,8) {$a$};
  \node [left] at (0,7) {$aa$};
  \node [left] at (0,6) {$aab$};
  \node [left] at (0,5) {$b$};
  \node [left] at (0,4) {$bb$};
  \node [left] at (0,3) {$c$};
  \node [left] at (0,2) {$ca$};
  \node [left] at (0,1) {$cb$};

  \node [right] at (2,9) {$\epsilon$};
  \node [right] at (2,8) {$b$};
  \node [right] at (2,7) {$ab$};
  \node [right] at (2,6) {$aab$};
  \node [right] at (2,5) {$bb$};
  \node [right] at (2,4) {$a$};
  \node [right] at (2,3) {$c$};
  \node [right] at (2,2) {$ca$};
  \node [right] at (2,1) {$cb$};

  \def\xa{-0.125}
  \def\xb{2.125}
  \draw [dotted,thick,gray] (\xa,9) to (\xb,9);
  \draw [dotted,thick,gray] (\xa,9) to (\xb,8);
  \draw [dotted,thick,gray] (\xa,9) to (\xb,5);
  \draw [dotted,thick,gray] (\xa,9) to (\xb,3);
  \draw [dotted,thick,gray] (\xa,9) to (\xb,2);
  \draw [dotted,thick,gray] (\xa,9) to (\xb,1);
  \draw [dotted,thick,gray] (\xa,5) to (\xb,9);
  \draw [dotted,thick,gray] (\xa,4) to (\xb,9);
  \draw [dotted,thick,gray] (\xa,3) to (\xb,9);
  \draw [dotted,thick,gray] (\xa,2) to (\xb,9);
  \draw [dotted,thick,gray] (\xa,1) to (\xb,9);

  \draw [thick] (\xa,5) to (\xb,8);
  \draw [thick] (\xa,3) to (\xb,8);
  \draw [red,very thick] (\xa,3) to (\xb,4);
  \draw [red,very thick] (\xa,9) to (\xb,6);
  \draw [red,very thick] (\xa,8) to (\xb,7);
  \draw [red,very thick] (\xa,7) to (\xb,8);
  \draw [red,very thick] (\xa,6) to (\xb,9);

  \node [right] at (-1.8, -0.5) { \textcolor{red}{$\mathcal{P} = \{\epsilon, a, aa, aab, c \}$} };
  \node [right] at (-1.8, -1.5) { \textcolor{red}{$\mathcal{S} = \{\epsilon, b, ab, aab, a \}$} };

  \def\xc{90}
  \matrix (m)
          [matrix of nodes,
            nodes in empty cells,
            anchor=base,            
          ] at (13,0) {
              & \node[rotate=\xc]{$\epsilon$}; & \node[rotate=\xc]{$b$}; & \node[rotate=\xc]{$ab$}; & \node[rotate=\xc]{$aab$}; & \node[rotate=\xc]{$bb$}; & \node[rotate=\xc]{$a$}; & \node[rotate=\xc]{$c$}; & \node[rotate=\xc]{$ca$}; & \node[rotate=\xc]{$cb$};\\
            $\epsilon$ & 1 & 1 & 0 & 1 & 1 & 0 & 1 & 1 & 1 \\
            $a$        & 0 & 0 & 1 & 0 & 0 & 0 & 0 & 0 & 0 \\
            $aa$       & 0 & 1 & 0 & 0 & 0 & 0 & 0 & 0 & 0 \\
            $aab$      & 1 & 0 & 0 & 0 & 0 & 0 & 0 & 0 & 0 \\
            $b$        & 1 & 1 & 0 & 0 & 0 & 0 & 0 & 0 & 0 \\
            $bb$       & 1 & 0 & 0 & 0 & 0 & 0 & 0 & 0 & 0 \\
            $c$        & 1 & 1 & 0 & 0 & 0 & 1 & 0 & 0 & 0 \\
            $ca$       & 1 & 0 & 0 & 0 & 0 & 0 & 0 & 0 & 0 \\
            $cb$       & 1 & 0 & 0 & 0 & 0 & 0 & 0 & 0 & 0 \\
          };
          
          \draw [thick] (7.6,9.8) rectangle (18.8,-.3);
          \draw [red,very thick] (7.4,10) rectangle (13.3,4.5);
          \draw [red,very thick] (7.4,2.5) rectangle (13.3,1.7);
          \draw [red,very thick] (14.6,2.5) rectangle (15.4,1.7);
          \draw [red,very thick] (14.6,10) rectangle (15.4,4.5);
          \node [right] at (7.7, 10.5) { \textcolor{red}{$\mathbf{H}^{\mathcal{P} \times \mathcal{S}}$} };
          \node [right] at (16, 10.3) { $\mathbf{H}^{\mathcal{P}_{\mathcal{T}} \times \mathcal{S}_{\mathcal{T}}}$ };
          
          \node [right] at (5.5, -1.5) { $\mathrm{rank}(\mathbf{H}^{\mathcal{P}_{\mathcal{T}} \times \mathcal{S}_{\mathcal{T}}}) = \mathrm{struct\_rank}(\mathbf{H}^{\mathcal{P}_{\mathcal{T}} \times \mathcal{S}_{\mathcal{T}}}) = 5$ };
           \node [right] at (5.5, -2.5) { $\mathrm{rank}(\textcolor{red}{\mathbf{H}^{\mathcal{P} \times \mathcal{S}}}) = \mathrm{struct\_rank}(\textcolor{red}{\mathbf{H}^{\mathcal{P} \times \mathcal{S}}}) = 5$ };

\draw [gray,thick,dashed] (-2.5,-1.5) to (-2.5,10);
\draw [gray,thick,dashed] (5,-1.5) to (5,10);


\def\xd{-16}


\node [right] at (\xd, 9) { $\Sigma = \{a, b, c\}$};
\node [right] at (\xd, 7.2) {

  $\begin{aligned}
    \mathcal{T} = \{ & (\epsilon,1), (aab,1),  (b,1),  (bb,1),  \\
    & (c,1), (ca,1), (cb,1) \}
  \end{aligned}$
};
\node [right] at (\xd, 5) {
  $f_{\mathcal{T}}(x) = \begin{cases}
    1 & \text{if}\ x \in \{\epsilon, aab, b, bb, c, ca, cb\} \\
    0 & \text{otherwise}
  \end{cases}
  $};
\node [right] at (\xd, 3) { $\mathcal{P}_{\mathcal{T}} = \{\epsilon, a, aa, aab, b, bb, c, ca, cb\}$};
\node [right] at (\xd, 2) { $\mathcal{S}_{\mathcal{T}} = \{\epsilon, b, ab, aab, bb, a, c, ca, cb\}$};
  
\end{tikzpicture}
  }
  \caption{Illustration of the Maximum Bipartite Matching
    sub-block. Left: a training set and the associated target
    function. Middle: a prefix-suffix graph with a corresponding
    maximum matching in red. Right: the full Hankel matrix for the
    training set, and the submatrix given by the  matching.}
  \label{fig:bmsub}
\end{figure*}

\section{PRELIMINARIES}



\subsection{Non Deterministic Weighted Finite State Automata}

We start by defining a class of functions over discrete
sequences. More specifically, let $x= x_1 \cdots x_t$ be a sequence
of length $t$ over some finite alphabet $\Sigma$. We use
$\Sigma^\star$ to denote the set of all finite sequences with elements
in $\Sigma$, and we use $\epsilon$ to denote the empty sequence.
The domain of our functions is $\Sigma^\star$. 

An Non-Deterministic Weighted Automaton (WA) with $n$ states is
defined as a tuple:
$A = \langle \azero, \ainf, \{ \A_{\sigma} \}_{\sigma \in \Sigma} \rangle$
where $\azero$, $\ainf \in \mathbb{R}^n$ are the initial and
final weight vectors and $A_{\sigma} \in \mathbb{R}^{n \times n}$ are
the transition matrices associated to each symbol $\sigma \in \Sigma$.
The function $f_{A}: \Sigma^{\star} \to \rr $ realized by an WA
$A$ is defined as:
\begin{equation}
f_A(x)=\azero^\top \A_{x_1}  \cdots \A_{x_t} \ainf \qquad . 
\end{equation}

The above equation is an algebraic representation of the computation
performed by an WA on a sequence $x$. To see this consider a state
vector $\mat{s}_i \in \rr^n$ where the $j$th entry represents the sum
of the weights of all the state paths that generate the prefix
$x_{1:i}$ and end in state $j$. Initially, $\mat{s}_0 = \azero$, and
then $\mat{s}_{i}^\top= \mat{s}_{i-1}^\top \A_{x_i}$ updates the state
distribution by simultaneously emiting the symbol $x_i$ and
transitioning to generate the next state vector.  WAs constitute a
rich function class which properly includes popular sequence models 
such as HMMs.

\subsection{Hankel Matrices}

We now introduce the concept of Hankel matrices for WA, which are
central to the spectral learning algorithm, and to the result in this
paper.

Let $f: \Sigma^\star \to \rr$ be an arbitrary function from sequences
to reals (not necessarily computed by a WA). Let $\Ps, \Ss \subseteq
\Sigma^\star$ be sets of sequences. We call prefixes the elements $p
\in \Ps$, and suffixes the elements $s \in \Ss$. The Hankel matrix $\H_f
\in \rr^{\Ps \times \Ss}$ for $f$ over the block $(\Ps, \Ss)$ is
defined by entries
$\H(p, s) = f(ps)$, 
where $ps$ is the concatenation of prefix $p \in \Ps$ and suffix $s
\in \Ss$. The following theorem gives a bijection between the class of
functions computed by WA and Hankel matrices:

\vskip 2em

\begin{theorem} \citep{schut61,CarlylePaz71,Fliess74}
  A function $f: \Sigma^\star \to \rr$ can be realized by a WA with
  $n$ states if and only if, for every possible block $(\Ps, \Ss)$,
  the corresponding Hankel matrix $\H_f$ has rank at most $n$.
\end{theorem}

\subsection{The Spectral Method}
\label{section:spectral}

We now give a brief description of the spectral method for estimating
a minimal WA representation for a target function. The algorithm is a
constructive version of the theorem above: it builds a Hankel matrix
of rank $n$ and computes the associated $n$ state WA from it.
We only provide a higher-level description of the method; for a
complete derivation and the theory justifying the algorithm we refer
the reader to the works by \citet{hsu09} and \citet{mlj12}.

Assume a training set $\Train$ in the form of a collection of
sequences, each paired with a target real value. We will denote as
$f_\Train$ the function obtained from the training set, i.e. if $x \in
\Train$, $f_\Train(x)$ is the target value.
For example, $\Train$ could be a corpus of English sentences, and
$f_\Train(x)$ the probability with which $x$ appears in $\Train$.

Given a training set $\Train$, the spectral algorithm computes a WA
$A$ with $n$ states, where $n$ is a parameter of the algorithm, such
that $f_A$ is a good approximation of $f_\Train$. See \cite{hsu09} for
the generalization theory of the algorithm.
The method is described by the following steps:
%
%
%
%
\begin{enumerate}[label=(\arabic*)]
\item Select a Hankel block. Let $\Ps_\Train$ and $\Ss_\Train$ be
  respectively the sets of all unique prefixes and suffixes of
  sequences in $\Train$. Select a block out of them, namely, a subset
  of prefixes $\Ps \subseteq \Ps_\Train$ and a subset of suffixes $\Ss
  \in \Ss_\Train$.
  
\item Compute Hankel matrices for $(\Ps,\Ss)$.
  \begin{enumerate}
  \item Compute $\H \in \rr^{\Ps \times \Ss}$, with entries
    \mbox{$\H(p,s) = f_\Train(ps)$}. 
  \item Compute $\mat{h}_\Ps \in \rr^\Ps$ with $\mat{h}_\Ps(p) =
    f_\Train(p)$ and $\mat{h}_\Ss \in \rr^\Ss$ with \mbox{$\mat{h}_\Ss(s) =
    f_\Train(s)$}.
  \item For each $\sigma \in \Sigma$, compute $\H_\sigma \in \rr^{\Ps \times \Ss}$ with entries
    \mbox{$\H_\sigma(p,s) = f_\Train(p \sigma s)$}.
  \end{enumerate}

\item Compute an $n$-rank factorization of $\H$. Compute the truncated SVD of $\H$, i.e. $\H \approx \mat{U} \mat{\Sigma}
  \mat{V}^\top$ resulting in a matrix $\mat{F} = \mat{U} \mat{\Sigma}
  \in \rr^{\Ps\times n}$ and a matrix $\mat{B}=\mat{V} \in \rr^{\Ss
    \times n}$.

  \item Recover the WA $A$ of $n$ states. Let $\mat{M}^+$ denote the
    Moore-Penrose pseudo-inverse of a matrix $\mat{M}$. The elements
    of $A$ are recovered as follows. Initial vector: $ \azero^\top= \mat{h}_\Ss^\top \mat{B}$. Final vector:  $ \ainf = \mat{F}^{+} \mat{h}_\Ps$. Transition Matrices: $\A_{\sigma} = \mat{F}^{+} \H_{\sigma} \mat{B}$, for $\sigma \in \Sigma$.
\end{enumerate}

There are some observations to make that motivate the contribution of
this paper. Consider the complete training block $(\Ps_\Train,
\Ss_\Train)$, and let $\H^\Train$ denote the Hankel matrix for this
complete block. If we want to fully reconstruct the function
$f_\Train$, we need an automata $A$ that has as many states as the
rank of $\H^\Train$. By using less states, we will be learning a
low-rank approximation of $f_\Train$ in the form of a WA.

The second observation is that any sub-block $(\Ps, \Ss)$ whose Hankel
submatrix has full rank (with respect to the rank of $\H^\Train$) can
be used to fully recover $f_\Train$.



Thus, in the ideal case, step (1) of the algorithm would select a
compact submatrix of $\H^\Train$ that preserves the rank.  By doing
so, the cost of steps (4) and (5) would only depend on the size of the
submatrix.
Even if we can not get the ideal block, it would be good to have a
method for step (1) that produces a small and informative block.
%
%
%
%
Unfortunately, in the general case (i.e. for any real matrix) finding
the submatrix of fixed size that has maximal rank is known to be
NP-complete \citep{subHard}.  In this paper we propose an algorithm to
find a small submatrix of $\H^\Train$ of high rank.
%



As a final note, spectral methods can be used to learn a language
model, that is, a probability distribution over all sentences of a
language. A straightforward way to learn a language model is
to regard the training collection $\Train$ as an empirical
distribution over sequences of words, where the probability of a
sequence is proportional to the number of times it appears,
i.e. $f_\Train(x) = \Pr_\Train(x)$. Another choice, sometimes referred
to as moment matching, is to set the function $f_\Train(x)$ to be the
expected number of times that the sequence $x$ appears as a
subsequence of a random sequence sampled from an empirical
distribution. In this case, the spectral algorithm will learn a WA
that computes expectations of subsequence frequencies. One useful
result is that this WA can be converted to another WA that
corresponds to the underlying language model, i.e. a distribution over
sequences; see \citet{mlj12} for details. In practice this second
method is preferred, since subsequence frequency expectations are
statistics that are more stable to estimate from a training set.

\section{SUB-BLOCK SELECTION VIA BEST BIPARTITE MATCHING}

We start this section by defining the structural rank of a matrix. 

\begin{definition}
The \textbf{structural rank} of a matrix is the maximum rank
of all numerical matrices with the same non-zero pattern. 
\end{definition}

Our
proposed algorithm will then search for a submatrix of $\H$ with full
structural rank.
In the
context of WA and Hankel matrices this has a nice interpretation as a
notion of \emph{complexity} of the support of a function. This is
because the structural rank of a Hankel matrix corresponds to the
number of states of the minimal WA for the \emph{hardest} function
defined over that support.

Notice that by definition, the numerical rank of a matrix is always
less or equal than its structural rank, thus the structural rank of
the Hankel matrix $\H$ of a function $f_A$ will be always greater or
equal than the number of states of the minimal WA computing
$f_A$. Our algorithm is based on finding a submatrix of $\H$ of full
structural rank.

The problem of finding a full structural rank sub-block of $\H$ can be
casted as an instance of maximum bipartite matching
\citep{edmonds67}. Given a bipartite graph $(V,G)$ where $V$ are the
set of vertices and $G$ the set of edges, the maximum bipartite
matching is defined as the largest set of non-intersecting
edges, where \emph{non-intersecting} means that no two edges in the
set share a common vertex.

In the case of the Hankel matrix for a function $f_A$ we would have a
bipartite graph $(V,G)$ where on one side we have vertices
corresponding to all unique prefixes in the support of $f_A$ and on
the other side we have all unique suffixes, thus: $|V|=|\Ps| +
|\Ss|$. There will be an edge connecting node $i$ an $ j$ if the
corresponding sequence made by the concatenation of prefix $i$ and
suffix $j$ is in the support of $f_A$. For every sequence $s$ of
length $T$ in the support of $f_A$ and every possible cut of $s$ into a
prefix and a suffix, there will be $T + 1$ corresponding edges in $G$,
thus $|G|=O(T|f_A|)$ where we use $|f_A|$ to refer to the number of
sequences in the support of $f_A$.

The maximum bipartite matching of a set of sequences is a subset of
the sequences such that no two sequences share a common prefix or
suffix and there is no larger subset that satisfies that property.
Figure \ref{fig:bmsub} shows an example of a function $f_A$ and its
corresponding graph, and a maximum bipartite matching for that graph.

We define the \emph{maximum bipartite matching sub-block} as the block
consisting of all vertices (prefixes and suffixes) in a maximum
matching. Figure \ref{fig:bmsub} shows an example of a function, a
maximum bipartite matching, and the corresponding sub-block and Hankel
submatrix.

To find a maximum bipartite matching there are several classical
algorithms. The Augmented Paths algorithm runs in $\BigO{|V||E|}$, but
in practice it has a much lower average case complexity. The
Hopcroft-Karp algorithm runs in $\BigO{|E| \sqrt{|V|}}$, removing the
linear dependence on $V$ (however, in our experiments the Augmented
Paths algorithm was alredy very fast). 
In the next section we propose an algoritm that takes advantage of the
structure of the Hankel matrix to obtain further speed ups. 


\subsection{On the Optimality of the Maximum Matching Sub-block}

\label{sec:theory}

We will use a weak version the \emph{matching property}, an assumption
used by \citet{hoffmanmccormick82}.
Let $\mat{M}$ be a matrix of structural rank $s$.
$\mat{M}$ has the weak matching property (WMP) if for any submatrix
$\mat{M}'$ of at least $s$ rows and $s$ columns, the rank of $\mat{M}'$
is equal to the structural rank of $\mat{M}'$.

\newcommand{\MBM}{\mathcal{B}}

\begin{lemma}  
  Let $\H$ be a Hankel matrix that satisfies the weak matching
  property.  Let $\MBM$ be a maximum bipartite matching of $\H$ and
  let $\H^\MBM$ be the corresponding submatrix.  $\MBM$ is a basis of
  $\H$, i.e. the rank of $\H^\MBM$ is equal to the rank of $\H$.
\end{lemma}

{\bf Proof.} Let $s(\mat{M})$ be the structural rank of a matrix
  $\mat{M}$. Let $n$ be the rank of $\H$, and note that $s(\H)$ is
   $n$ because $\H$ has WMP.
Now note that $s(\H^\MBM)$ is also $n$, because the maximum bipartite
matching of $\H$ is included in $\H^\MBM$, thus $s(\H^\MBM)$ is at
least $n$; and it is at most $n$, otherwise $s(\H) \ge s(\H^\MBM) > n$.
Since $\H$ has WMP, the rank of $\H^\MBM$ is $n$. \qed

Ideally, we would not have to assume the \emph{matching property} and
instead we could provide theoretical guarantees for the maximum gap
between the structural and numeric rank of a matrix. Unfortunately,
because of the discrete nature of the structural rank, deriving
useful bounds for this gap has been shown to be a hard theoretical
challenge \citep{hoffmanmccormick82}. Thus to provide validation for
our assumption we resorted to an empirical evaluation of the gap on a
wide range of sequence modeling datasets, where we observe that the weak matching property is a reasonable assumption. 
%
The complete results are in section B of the supplementary
materials.

\section{FASTER BIPARTITE MATCHING FOR HANKEL MATRICES}

As said in the previous section, finding the structural rank can be
reduced to the maximum bipartite matching problem. In this section, we
propose a simple heuristic to speed-up the maximum bipartite matching
for the specific case where the underlying matrix is a Hankel.  We do
this by exploiting structural properties of these matrices for an
underlying subroutine, the \textit{augmenting path} algorithm.  Each
basic application of the augmenting path increases the matching by
one, and a matching is maximum if and only if there is no further
augmenting path.  The straightforward solution of applying it on each
node is equivalent to the maximal flow algorithm, and while more
sophisticated algorithms where proposed~\citep{Hopcroft1973} which
find several paths per iteration, benchmarks~\citep{Setubal1996} have
shown that the simple algorithm works in general faster.

\def\myskip{0.2cm}

\begin{figure}[t]
  \begin{center}
  \begin{tikzpicture}[scale=0.7]
    \node at (1,1.5) {(a)};
    \node [circle, fill=black, inner sep=0.05cm, label=left:$x$] at (0,1) (x) {};
    \node [circle, fill=black, inner sep=0.05cm, label=left:$y$] at (0,0) (y) {}; 
    \node [circle, fill=black, inner sep=0.05cm, label=right:$z$] at (2,1) (z) {};
    \node [circle, fill=black, inner sep=0.05cm, label=right:$w$] at (2,0) (w) {}; 
    \draw [thick] (x) to (z);
    \draw [thick] (y) to (z);
    \draw [thick] (y) to (w);     
  \end{tikzpicture}
  \hskip \myskip
  \begin{tikzpicture}[scale=0.7]
    \node at (1,1.5) {(b)};
    \node [circle, fill=black, inner sep=0.05cm, label=left:$x$] at (0,1) (x) {};
    \node [circle, fill=black, inner sep=0.05cm, label=left:$y$] at (0,0) (y) {}; 
    \node [circle, fill=black, inner sep=0.05cm, label=right:$z$] at (2,1) (z) {};
    \node [circle, fill=black, inner sep=0.05cm, label=right:$w$] at (2,0) (w) {}; 
    \draw [arrows={-latex},thick] (x) to (z);
    \draw [arrows={latex-},thick] (y) to (z);
    \draw [arrows={-latex},thick] (y) to (w);     
  \end{tikzpicture}
 \hskip \myskip
  \begin{tikzpicture}[scale=0.7]
    \node at (1,1.5) {(c)};
    \node [circle, fill=black, inner sep=0.05cm, label=left:$x$] at (0,1) (x) {};
    \node [circle, fill=black, inner sep=0.05cm, label=left:$y$] at (0,0) (y) {}; 
    \node [circle, fill=black, inner sep=0.05cm, label=right:$z$] at (2,1) (z) {};
    \node [circle, fill=black, inner sep=0.05cm, label=right:$w$] at (2,0) (w) {}; 
    \draw [arrows={latex-},thick] (x) to (z);
    \draw [arrows={-latex},thick] (y) to (z);
    \draw [arrows={latex-},thick] (y) to (w);     
  \end{tikzpicture}
  \end{center}
  \caption{Illustrations of the augmenting path algorithm.}
  \label{fig:aug}
\end{figure}
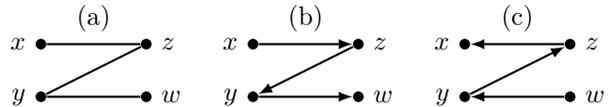

We first describe the basic procedure: assume the graph depicted in Figure \ref{fig:aug} (a), 
and furthermore assume that the current matching (not maximal) is as follows: $M = \{(y,z)\}$.
This is clearly not maximal, as a better (and maximal) matching would be $\{ (x,z), (y,w)\}$.
The augmenting path procedure maps the previous matching to the \textit{directed} graph $G$ depicted in Figure \ref{fig:aug} (b).
%
%
%

Unselected edges will be directed from left to right, while selected edges will be directed from right to left.  
An augmenting path is then defined as a path $x_1, \dots, x_m$ over $G$, such that $x_1$ belongs to the left partition, $x_m$ to the right one, and both $x_1$ and $x_m$ are unmatched (this is, do not belong yet to a matching).  
No restrictions are put on the intermediate nodes, but it becomes clear that the path alternates between unmatched pairs (left to right edges) and matched pairs (right to left).  
Note that such paths can now easily be retrieved with a standard graph traversal (in our implementation we use a depth-first search, which we assumed was faster on sparse graphs although this was not verified).  
Starting from node $x$, the following path can then be retrieved: $x,z,y,w$, and the graph will then be rewired to the graph depicted in Figure \ref{fig:aug} (c).
No further augmenting paths exist here, and the maximum matching algorithm therefore finishes with the following matching: $\{(x,z), (y,w)\}$.

The specific case where the left part of the bipartite are prefixes and the right part are suffixes creates some strong structural constraints.
Notably:

\begin{property}
$(p \sigma, s)$ is an edge in the graph iff $(p, \sigma s)$ is an edge
\label{prop:hankelBiGraphs}
\end{property}
This is, the edges of the bipartite graph denoting a Hankel matrix come by (possibly overlapping) groups of edges, each group originating in one of the support sequences.

We propose to take advantage of that structural knowledge to speed-up the maximum matching algorithm.
First, we sort the prefixes by their lengths, and start applying the augmenting path procedure from the longest prefix node.
Each augmenting path procedure returns a set of edges $R$ to be removed from the matching, and a set of edges $A$ to be added to the matching.
For each edge $(\sigma_1 \dots \sigma_k, s) \in A$ we consider all \textit{shifted} pairs $(\sigma_1 \dots \sigma_i, \sigma_{i+1} \dots \sigma_k s)$.
Due to Property~\ref{prop:hankelBiGraphs}, each one of these pairs is an edge in the bipartite graph.
We check each such pair, and if both nodes are unmatched we simply add them to the matching.

Assuming a bitset implementation of sets, the checks can be done in $\BigO{|E|}$, but in the worst-case scenario, it may well be that none of the shifted pairs are free, and therefore only add computation without improving the matching.
In \S~B of the supplementary material we report synthetic
experiments that show the speed-ups of this strategy compared to the
standard method.


\section{EXPERIMENTS}

To validate our sub-block selection strategy, we present 
comparisons to methods for scaling up spectral learning. We first
compare to general methods to scale SVD, and then to sub-block
selection strategies for Hankel matrices.
We end this section with a comparison to state-of-the-art methods on
the SPiCE benchmark. 

In all experiments we use natural language data for the task of
language modeling. The goal is to learn a language
model that predicts the next symbol for a sentence prefix
(including ending the sentence).
As evaluation metric we use \emph{Bits per Character (BpC)}, the
average log-2 probability that the model gives to each symbol
in the evaluation senquences, including sequence ends.
As datasets we use the English Penn Treebank \citep{penntreebank}
using standard splits\footnote{49 characters; 5017k / 393k / 442k
characters in the train / dev / test portions.},
the War and Peace dataset \citep{karpathy16}\footnote{84 symbols; 2658k / 300k / 300k
  characters in the train / dev / test portions.}, and the NLP
datasets of the SPiCe benchmark \citep{spice}.

\subsection{Scalable SVD Methods}


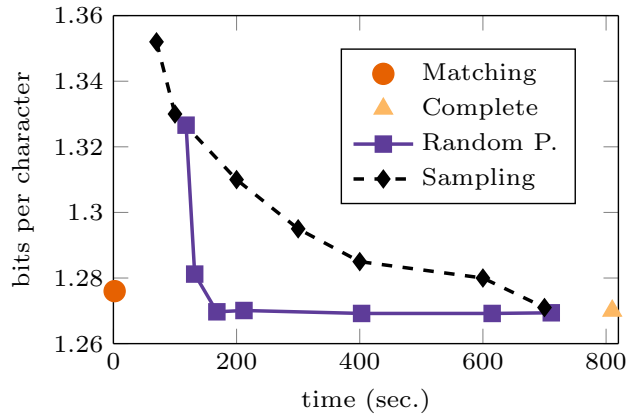
\begin{figure}[t]
  \definecolor{color1}{RGB}{230,97,1}
  \definecolor{color2}{RGB}{253,184,99}
  \definecolor{color3}{RGB}{1,1,1}
  \definecolor{color4}{RGB}{94,60,153}
  \begin{center}
    \resizebox{0.5\textwidth}{!}{
    \begin{tikzpicture}
      \scriptsize
      \begin{axis}[%
          width=2in,
          height=1.3in,
          at={(1.44217in,0.915223in)},
          scale only axis,
          separate axis lines,
          every outer x axis line/.append style={black},
          every x tick label/.append style={font=\color{black}},
          xmin=0,
          xmax=820,
          xlabel near ticks,
          ylabel near ticks,
          xlabel={time (sec.)},
          every outer y axis line/.append style={black},
          every y tick label/.append style={font=\color{black}},
          ymin=1.26,
          ymax=1.36,
          ylabel={bits per character},
          every node near coord/.append style=={font=\tiny},
          clip=false,
          legend style={at={(0.45,0.9)},anchor=north west,legend cell align=left,align=left,draw=black},
        ]
        \addplot [color=color1,only marks, mark size=3.0pt, mark=*,mark options={solid}]
        table[row sep=crcr]{%
          2   1.2760\\
        };
        \addlegendentry{Matching};
        \addplot [color=color2, only marks, mark size=3.0pt, mark=triangle*, mark options={solid}]
        table[row sep=crcr]{%
          810  1.2700\\
        };
        \addlegendentry{Complete};
        \addplot [color=color4, solid, line width=1.0pt,mark size=2.0pt, mark=square*, mark options={solid}]
        table[row sep=crcr]{%
          118     1.3266\\
          132     1.2812\\
          168     1.2697\\
          212     1.2701\\
          403     1.2692\\
          615     1.2692\\
          711     1.2694\\
        };
        \addlegendentry{Random P.};
        \addplot [color=color3, dashed,line width=1.0pt,mark size=2.0pt,mark=diamond*,mark options={solid}]
        table[row sep=crcr]{%
          70    1.3520\\
          100   1.3300\\
          200   1.3100\\
          300   1.2950\\
          400   1.2850\\
          600   1.2800\\
          700   1.2710\\
        };
        \addlegendentry{Sampling};
            
      \end{axis}
    \end{tikzpicture}%
    }
  \end{center}
  \vspace*{-0.6cm}
  \caption{Comparison of Strategies for Scaling Spectral Learning.}
  \label{timetradeoff}
\end{figure}

We conducted experiments comparing our method with two other
strategies for scaling SVD. The first uses \emph{Randomized
  Projections} to perform SVD \citep{rp_1}. This idea was previously
used to scale spectral learning \citep{ICML2013_hamilton13}.  The second strategy is
based on \emph{Sampling}, and selects the $k$ top rows and columns
that have the highest norm \citep{sampling}.

For this comparison we used the Penn TreeBank dataset with simplified
part-of-speech tags (12 symbols). We chose
this dataset because it results in a relatively small Hankel matrix
where we can run sparse SVD. In particular, we used moment size of
$T=5$, which results in a square Hankel matrix of size 52,450, numeric
rank of 312, and structural rank of 313. Thus, the \emph{Complete}
method will use run sparse SVD on this matrix. 

We present a trade-off between performance (in terms of
bits-per-character) and training time of a method. When appropriate,
we generate solutions that utilize different amounts of time. For
Sampling, since it selects $k$ rows and columns proportional to their
norm, a natural way of generating different solutions is to vary
$k$. For Randomized Projections we do not select a sub-block, instead
we project the Hankel matrix to a lower $\ell$-dimensional space and then
the SVD is performed on the projected matrix. Thus to get performance
as a function of training cost we can change the size of the
projection.

The training time\footnote{All experiments were run on a 2.2 GHz Intel Core processor.} of a method consists of: (1) time spent in selecting
the Hankel sub-block (for algorithms that start by sub-block selection
(e.g. best matching); (2) time spent on computing the singular value
decomposition; and (3) time spent computing inverses, i.e. recovering
operators. Notice that all spectral methods will perform SVD of a Hankel
sub-block. Whenever we compute SVD we take the cost of the most
efficient (i.e. sparse or full SVD) to be the cost of the
algorithm. Another important observation is that the sparse SVD
algorithm takes as a parameter the number of singular values to
compute. We take this to be the optimal number of states found using
the validation data.

Figure \ref{timetradeoff} shows the trade-off, for the four methods. 
The first observation is that with sufficient amount of computational
time both Random Projections and Sampling achieve the same performance
as using the Complete Hankel. This is expected since by setting $k$
and $\ell$ sufficiently large we should always obtain the same result as
using the complete Hankel. Random Projections seems to be
significantly better than Sampling in terms of speed up, and it can
obtain the same solution as Complete in less than $1/4$ of the
time. Best Bipartite Matching obtains a slightly higher
bits-per-character than Random Projections, but is significantly more
efficient. More precisely, to achieve the same performance as with
Matching, Random Projections requires about 50 times more time.

\subsection{Sub-block Selection Strategies for Spectral Methods}


We now present an empirical comparison between the most prevalent
sub-block selection strategies for spectral learning.

We train spectral language models at character level that use a fixed window of $T$
characters both at training and test time.
At training, we collect all substrings $x$ of length up to
$T$. Following \cite{mlj12}, we set a target function $f_\Train(x)$ to
be the expected number of times that $x$ appears as a subsequence of a
random sentence sampled from training.  We run the spectral algorithm
with $f_\Train$ and obtain a WA. At test, we run the WA to compute
the probability of the next character given a sliding prefix of length $T-1$.

We compare maximum matching sub-block selection to three strategies:
full block, random cuts, and length up-to.
\emph{Full block} uses all substrings of the support of $f_\Train$ as
prefixes and suffixes.
\emph{Random Cuts} follows \citet{balle12}: it samples a string $x$ of the
support, and chooses a random cut of $x$ into a prefix and suffix,
which are added to the sub-block. This process is repeated until the
sub-block reaches size $k$ (a parameter).
\emph{Length $\le \ell$} selects all substrings up to length $\ell$.

%


\begin{table}

  \def\SSS{Random Cuts }
   \def\LLL{Length $\le$ }
  \caption{Comparison Between Sub-block Selection Methods for Support Strings up to Size $T=5$}
  \label{tab:results}
  \begin{center}
  \begin{tabular}{@{}lrrrr@{}} \toprule
     {\bf method} & {\bf size} & {\bf rank} & {\bf sec.} & {\bf BpC} \\
     \midrule
     Full                      &   144,378    & -          & 18,000       & 1.735 \\
     Matching                      & 1,661    & 1,612      & 8       & 1.741 \\
     \SSS $1\times$          & 1,661    & 739        & 10      & 2.011 \\
     \SSS  $2\times$          & 3,322    & 807        & 74      & 1.828 \\
     \SSS  $3\times$          & 4,983    & 902        & 163     & 1.812 \\
     \SSS  $4\times$          & 6,664    & 989        & 271     & 1.791 \\
     \SSS  $5\times$          & 8,305    & 1,010      & 302     & 1.769 \\
     \SSS  $6\times$          & 9,966    & 1,086      & 411     & 1.761 \\
     \SSS  $7\times$          & 11,627   & 1,114      & 825     & 1.752 \\
     \LLL   2                 & 861      & 92         & 2       & 3.105 \\
     \LLL   3                 & 7,455    & 417        & 290     & 2.662 \\
     \LLL   4                 & 38,314   & 907        & 3,500   & 1.856 \\
     \bottomrule
   \end{tabular}
  \end{center}
\end{table}


Table \ref{tab:results} compares sub-block selection methods in
terms of the numeric rank of the sub-matrix, the time it takes to
compute an $n$-rank factorization, and the quality of the resulting
$n$-state WA in terms of bits per character (BpC). $n$ is a parameter
that we tune on validation data with a range of values up to the rank of the
sub-matrix.
The matching sub-block obtains results that are very close to using
the full matrix. However, it is much faster: the time to compute a
matching is negligible, and the time to factorize the matrix is three
orders of magnitude faster.
Compared to other strategies, the matching sub-block is the most
accurate and the most compact, and thus it is drastically faster. 
This improvement is achieved because it selects a very compact
sub-block (of size 1,661) that has approximately full numeric rank
(rank is 1,612).
In contrast, using Random Cuts, a block of the same size as the
matching sub-block (1,661) has only a rank of 739, which results in
lower quality predictions. When increasing the block of Random Cuts up
to 7 times the size of the matching, we obtain a rank of 1,114 and
very close results to the matching and full sub-blocks; however,
factorizing the sub-matrix is 100 times more costly.
Sub-block selection by maximum length also performs poorly. 
This last result is evidence that long-range statistical dependencies
exist in this data, and these are not captured by small moments. On
the other hand, a brute-force approach to capture such long range
dependencies is prohibitive. Our method clearly offers a very
competitive solution.

  

\begin{table}
  \def\BPC{{\scriptsize (BpC)}}
  \def\LIK{{\scriptsize (Lik)}}
  
  \caption{Results of spectral models for increasing length of strings in the support}
  \label{tab:bpc}
  \begin{center}
  \begin{tabular}{@{}lcclcccc@{}} \toprule
     & \multicolumn{2}{c}{Penn Treebank} && \multicolumn{3}{c}{War and Peace \LIK} \\
    \cmidrule{2-3} \cmidrule{5-7}
    $T$   & Full  & Match. && Full & Match. & KJL16 \\
    \midrule
     5   & 1.735  &  1.741   &&   1.377    &  1.405  & 1.451 \\
     6   & 1.623  &  1.653   &&   1.326    &  1.393  & 1.339  \\
     7   & 1.597  &  1.622   &&   1.323    &  1.369  & 1.321  \\         
     \bottomrule
   \end{tabular}
  \end{center}
\end{table}

\begin{table*}[t]
 \caption{Results on NLP Datasets of SPiCe Sequence Prediction Competition.}
  \label{tab:resultsspice}
  \begin{center}
    \begin{tabular}{@{}lrrrrrr@{}}
      \toprule
                    & Verbs   & LM (words)      & LM (characters) &  POS & Normalization & All \\ 
      \midrule 
   RNN-P      & \bf{*0.6078*} & \bf{*0.5434*}   & \bf{*0.8101*} & \bf{*0.6573*}   & \bf{*0.5882*} &\bf{*0.6414*} \\
   COMBO-NN-1 & 0.5794        & \bf{0.5014}          & 0.7632        & 0.6331          & 0.5181        &0.5990     \\
   COMBO-B    & 0.5514        & 0.4264          & \bf{0.7978}   & 0.5890          & 0.3843        &0.5498     \\
   LSTM       & 0.5123        & 0.4034          & 0.7630 &      0.5941            & 0.4187        &0.5383     \\
   COMBO-Sp   & 0.5273        & 0.4148          & 0.6142        & 0.6235          & 0.4990        &0.5358     \\
   Sp-BM      & \bf{0.5928}   & 0.4998          & 0.7820        & \bf{0.6356}     & \bf{0.5441}   &\bf{0.6109} \\
       \bottomrule
    \end{tabular}
  \end{center}
\end{table*}


Next we present results of models trained on larger substrings, of up
to size 7, for the Penn Treebank and War and Peace datasets. Table
\ref{tab:bpc} compares the performance of the matching sub-block to
using the full block.\footnote{For the War and Peace data, we measure
  performance in terms of test negative log-likelihood, such that we
  can compare to published results.} As we increase the size of the
substrings ($T$) the models get better. There is always a performance
gap between using the full or the matching blocks, however the
matching sub-block scales much better: the cost of computing a
matching is negligible (less than 15 seconds), and the cost of the
factorization is at three orders of magnitude faster.
Table \ref{tab:bpc} also compares to the results by \citet{karpathy16}
(noted KJL16), in terms of negative log-likelihood on test characters
(noted Lik). We report their results, corresponding to non-recurrent
feed-forward neural models, which condition each prediction on the
$T-1$ latest characters (see Table 2 of their paper). The
results are fairly comparable, exhibiting the same trend.

\subsection{Comparison with State-of-the-Art}

In order to compare the performance of our proposed method to other
state of the art methods for sparse sequence modeling, we run
experiments on the five NLP datasets of the SPiCe sequence prediction
competition \citep{spice}. The task of the competition was the following: given a
string (prefix) of symbols in a finite alphabet the goal is to predict
a ranking of possible next symbols to be the next element of the
sequence. The metric used for evaluation measures the average ranking
that the model gives to the correct next symbol.\footnote{We refer the
  reader to the SPiCe benchmark website:
  \url{http://spice.lif.univ-mrs.fr}.}  Both validation and test sets are
available from the challenge website.

There were a total of 26 teams implementing a wide range of methods,
including: many different types of neural network models, boosting,
spectral and classical state-merging algorithms for learning weighted
automaton, and ensemble methods that combined several techniques.

Table \ref{tab:resultsspice} shows results for the top 5 teams of the
competition. The top team (RNN-P) is a novel RNN architecture where the
state vector is augmented with an indicator vector representing the
previous ngram in the history. The second best team (COMBO-NN) is an
ensemble of MLP, CNN, LSTM and ngram models. The third team (COMBO-B)
is also an ensemble method of ngram, spectral, RNN and tree
boosting. The fourth team (LSTM) is an RNN with LSTM cells and the
fifth team (COMBO-Sp) is another ensemble method that combines a
spectral model with ngram models.

The performance of our spectral method with the proposed sub-block
selection using best bipartite matching (Sp-BM) is given in the last
row. We indicate with bold and stars the top performing method for
each dataset and with bold the second best. Running the proposed
algorithm out-of-the-box and without any model combination we get a
very competitive performance: second best overall (0.6414 vs 0.6109)
and second in 3 out of 5 datasets.  
One of the most attractive properties of our method is that
the most costly training times (those corresponding to datasets with
Hankel matrices of higher structural rank) were less than 5 minutes.


\section{CONCLUSIONS}

We presented a novel strategy for scaling spectral
learning algorithms that is specifically designed for modeling long
range dependencies in sparse sequence functions. The main idea is to
use maximal bipartite matching  to find a Hankel sub-block
of maximal structural rank. Our experiments on a real sequence
modeling task show that: (1) Exploiting large Hankel matrices is
essential for the success of spectral learning algorithms; and that:
(2) Our proposed sub-block selection strategy to handle large Hankel
matrices can be much faster than using sparse SVD over the complete
Hankel matrix without a significant loss in performance. Our algorithm
leads to a very appealing trade-off between computational complexity
and model performance.

\bibliographystyle{plainnat}
\bibliography{refs}

\end{document}